\documentclass[conference]{IEEEtran}
\IEEEoverridecommandlockouts
\usepackage{cite}
\usepackage{amsmath,amssymb,amsfonts}
\usepackage{algorithm}
\usepackage{algorithmic}
\usepackage{graphicx}
\usepackage{textcomp}
\usepackage{xcolor}
\usepackage{float}
\usepackage{multirow}
\usepackage{amssymb}
\usepackage{bm}

\usepackage{url}

\def\BibTeX{{\rm B\kern-.05em{\sc i\kern-.025em b}\kern-.08em
    T\kern-.1667em\lower.7ex\hbox{E}\kern-.125emX}}

\makeatletter
\newcommand\bigDiamond{\mathop{\mathpalette\bigDi@mond\relax}}
\newcommand\bigDi@mond[2]{%
  \vcenter{\hbox{\m@th
    \scalebox{\ifx#1\displaystyle 2\else1.2\fi}{$#1\Diamond$}%
  }}%
}
\newcommand\bigLozenge{\mathop{\mathpalette\bigL@zenge\relax}}
\newcommand\bigL@zenge[2]{%
  \vcenter{\hbox{\m@th
    \scalebox{\ifx#1\displaystyle 2\else1.2\fi}{$#1\blacklozenge$}%
  }}%
}
\makeatother

\definecolor{color_yuqi}{RGB}{39, 155, 228}

\begin{document}

\title{Transmission Line Outage Probability Prediction Under Extreme Events Using Peter-Clark Bayesian Structural Learning\\
{\footnotesize }
\thanks{This research work is supported by the U.S. National Science Foundation (NSF) Colorado-Wyoming Climate Resilience Engine program.}
}

\author{\IEEEauthorblockN{Xiaolin Chen}
\IEEEauthorblockA{\textit{Electrical Engineering Department} \\
\textit{Colorado School of Mines}\\
Golden, CO, USA \\
xiaolin.chen@mines.edu}
\and
\IEEEauthorblockN{Qiuhua Huang}
\IEEEauthorblockA{\textit{Electrical Engineering Department} \\
\textit{Colorado School of Mines}\\
Golden, CO, USA \\
qiuhuahuang@mines.edu}
\and
\IEEEauthorblockN{Yuqi Zhou}
\IEEEauthorblockA{\textit{Power Systems Engineering Center} \\
\textit{National Renewable Energy Laboratory}\\
Golden, CO, USA\\
yuqi.zhou@nrel.gov}
}

\maketitle

\begin{abstract}
Recent years have seen a notable increase in the frequency and intensity of extreme weather events. With a rising number of power outages caused by these events, accurate prediction of power line outages is essential for safe and reliable operation of power grids. The Bayesian network is a probabilistic model that is very effective for predicting line outages under weather-related uncertainties. However, most existing studies in this area offer general risk assessments, but fall short of providing specific outage probabilities. In this work, we introduce a novel approach for predicting transmission line outage probabilities using a Bayesian network combined with Peter-Clark (PC) structural learning. Our approach not only enables precise outage probability calculations, but also demonstrates better scalability and robust performance, even with limited data. Case studies using data from BPA and NOAA show the effectiveness of this approach, while comparisons with several existing methods further highlight its advantages.
\end{abstract}

\begin{IEEEkeywords}
Bayesian network, PC algorithm, structural learning, outage probability, extreme weather events.
\end{IEEEkeywords}

\section{Introduction}

With climate change intensifying globally, extreme weather events become more frequent and impactful than ever before. As of November 1, 2024, the United States has been hit with 24 separate billion-dollar disasters\cite{CiteNOAAdisaster}, which include severe weather events such as storms, cyclones, and wildfires. 
% Extreme climate and weather events over the past few years have led to a significant increase in power device outages and elevated risks of power systems, impacting the reliable operation and overall resilience of the entire systems \cite{ke2022data}. 
As a result, there has been a substantial increase in power device outages, which poses risks to the reliability of power systems \cite{ke2022data}. For example, extremely high temperatures and heat waves limit the transmission line capability, leading to line sagging and even causing high impedance faults \cite{zhou2024data}. Storms and hurricanes can cause faults and damage power lines. Cold waves can lead to failures in overhead lines and towers \cite{panteli2015influence}, including outdoor transformers and switchgear.

Although there have been existing studies that explore weather-related outage prediction for power devices, most approaches provide only general risk levels for power equipment, not specific outage probabilities \cite{yang2020enhancing}, which limits their applicability and effectiveness for system-level reliability and resilience analysis. The prediction of weather-related outages is often based on data-driven modeling techniques. For example, a logistic regression model is adopted for operational reliability prediction considering multiple meteorological factors in \cite{chen2018data}. Fragility curves deduced from weather-related outage databases are used to estimate component failure probability in \cite{ke2022data}. The application of the CatBoost algorithm, random forest, support vector machine, and neural network methods for predicting outages under severe weather conditions has been also explored in \cite{baembitov2024sensitivity}. Other research considers weather-dependent outage mechanisms and proposes weather outage rate models that take into account line conductor temperatures \cite{yao2018toward}, and wind speed and lightning\cite{alvehag2010reliability}. However, these methods need a large outage dataset to perform well, but such datasets are typically limited in real-world scenarios.

To address the aforementioned issues of the existing approaches, in this paper, we propose a novel transmission lines outage probability prediction method using a Peter-Clark (PC) structural learning-based Bayesian network. The major contributions of this paper include the following:

1) The proposed method leverages data-driven machine learning techniques to analyze the correlations among various meteorological factors and their impact on transmission line outages. Instead of merely providing several risk levels, it enables the precise calculation of outage probabilities.

2) To the best of our knowledge, the PC algorithm is applied for the first time in the reliability prediction of transmission lines for improved efficiency and scalability.

3) The proposed method works effectively even with limited outage data and outperforms other existing methods, which highlights its robustness.

The rest of this paper is organized as follows. The fundamentals and basic concepts of the Bayesian network are first introduced in Section~\ref{Sec: BNP}. After that, the PC algorithm for structural learning of the Bayesian network is presented in Section~\ref{Sec: PC}. In Section~\ref{Sec: Case}, we use case studies with realistic data to illustrate how the algorithm can be applied to predict transmission line outage probabilities. Lastly, the conclusions are presented in Section~\ref{Sec: conclu}.

\section{Bayesian Network Prediction}
\label{Sec: BNP}

We start by covering some fundamentals and basic concepts of Bayesian network prediction to better explain how the PC algorithm can be applied to improve efficiency and scalability in Bayesian network structure learning. These concepts include Bayes’ theorem, conditional independence, and Bayesian network, which will later help us illustrate the PC structure learning process.

\subsection{Bayes' Theorem}
\label{Sec: Bayes}
Bayes' theorem \cite{pishro2014introduction} gives a mathematical rule for inverting conditional probabilities, which allows us to find the probability of a cause given an observed effect. The theorem can be mathematically expressed as:
\begin{equation}
P(A \mid B) =\frac{P(B \mid A) P(A)}{P(B)}
\label{eq:bayes}
\end{equation}
The likelihood function \( \textstyle P(B \mid A) \) represents the probability of the evidence \( B \) given that \( A \) is true, the posterior probability \( \textstyle P(A \mid B) \) represents the probability of \( A \) after considering the evidence \( B \), \( P(A) \) is the prior probability, and \( P(B) \) is the probability of the evidence.

\subsection{Conditional Independence}
\label{Sec: CI}
Conditional independence is another important concept in causal inference analysis \cite{dawid1980conditional}. It states that two events $A$ and $B$, are conditionally independent given an event $C$ with $P(C)>0$ if 
\begin{equation}
P(A \cap B \mid C) = P(A \mid C) P(B \mid C)
\label{eq:condition inde}
\end{equation}
Hence, when given $C$, the occurrence of $A$ does not provide any additional information about the occurrence of $B$.

\subsection{Bayesian Network}
\label{Sec: BN}
\begin{figure}[t!]
\vspace{2mm}
\centering
\includegraphics[trim=0cm 0cm 0cm 0cm,clip=true,totalheight=0.075\textheight]{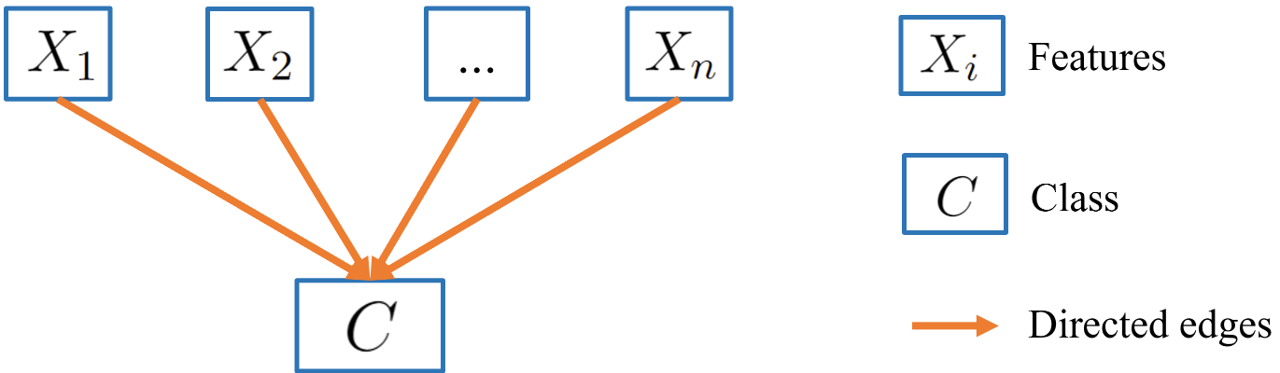}
\caption{The general structure of a Naive Bayes classifier.}
\label{fig:naive}
\vspace{-1mm}
\end{figure}

A Bayesian network is a probabilistic graphical model that represents a set of random variables and their conditional dependencies via a directed acyclic graph (DAG). 
% There are two main components in a Bayesian network: a DAG, and a joint probability distribution.
Among all Bayesian networks, the Naive Bayes classifier is a specific type of Bayesian network
% that assumes the value of each feature is independent of the value of any other feature, given the class variable.
in which each feature is directly connected to the class variable but is independent of the other features (see Fig.~\ref{fig:naive}). This structure simplifies the model to a single-layer network, where the class node is the parent of all feature nodes, making it a ``naive'' simplification of a Bayesian network. Accordingly, the Naive Bayes classifier uses Bayes' Theorem to calculate the posterior probability of a class $k \in \cal{K}$ given its features:
\begin{equation}
P(C_k \mid \bm{X}=\bm{x}) = \frac{P(C_k) \prod_{i=1}^n P(X_i=x_i \mid C_k)}{P(\bm{X}=\bm{x})}  
\label{eq:naive}
\end{equation}
in which \( \textstyle P(C_k \mid \bm{X} = \bm{x}) \) denotes the probability of class $C_k$ given the feature vector $\bm{x}$; $P(C_{k})$ denotes the prior probability of class $C_{k}$; $\prod_{i=1}^n P(X_i=x_i \mid C_k)$ represents the product of all features  based on the Naive Bayes assumption of conditional independence; 
$P(\bm{X}=\bm{x})$ is the probability of the feature vector $\bm{x}$.

Considering the Markov property of Bayesian network, the factorization of the Bayesian network can be represented as follows:

\begin{equation}
P(X_1,...,X_n) = \prod_{i=1}^n P\left(X_i \mid \text{Pa}(X_i)\right) 
\label{eq:factorization}
\end{equation}
in which $X_i$ are the variables of the Bayesian network, and $\text{Pa}(X_i)$ is the set of parent nodes of variable $X_i$.

% \begin{figure}[t!]
% \centerline{\includegraphics[width=7cm]{Naive bayesian5.png}}
% \caption{General structure of a Naive Bayesian network}
% \label{fig:naive}
% \vspace{-3mm}
% \end{figure}

An advantage of Naive Bayes is that it requires only a small amount of training data to estimate the parameters necessary for classification. 
% And no iterative process is needed as the probabilities can be directly computed, which helps alleviate problems stemming from the curse of dimensionality, such as the need for data sets that scale exponentially with the number of features.
Since the probabilities can be computed directly, no iterative process is required, which helps to reduce problems caused by the curse of dimensionality.
However, Naive Bayes often fails to produce a good estimate for the correct class probabilities. 
% Even though the output of a Naive Bayes classifier is a probability score, it is not the true probability, but is proportional to the true probability. 
Although the output of a Naive Bayes classifier is a probability score, it represents a value proportional to the true probability rather than the true probability itself.

Moreover, the assumption of conditional independence among all features often does not hold, as features show some form of dependency in most situations.
Therefore, it is essential to apply causal discovery algorithms to determine the dependencies among various meteorological factors, which can improve the accuracy and scalability of the Bayesian network for line outage prediction tasks.

\section{PC Structural Learning}
\label{Sec: PC}

The Peter-Clark (PC) algorithm\cite{spirtes2001causation} is a powerful and efficient tool for learning the structure of Bayesian networks. 
The PC algorithm stands out for its efficiency when compared with other causal discovery algorithms, because it restricts the search space for conditional independence tests by progressively deleting irrelevant connections from an initially fully connected undirected graph. The algorithm only requires independence tests among directly connected nodes, and thus significantly reduces the total number of tests needed.

\begin{figure}[t!]
\vspace{2mm}
\centering
\includegraphics[trim=0cm 0cm 0cm 0cm,clip=true,totalheight=0.12\textheight]{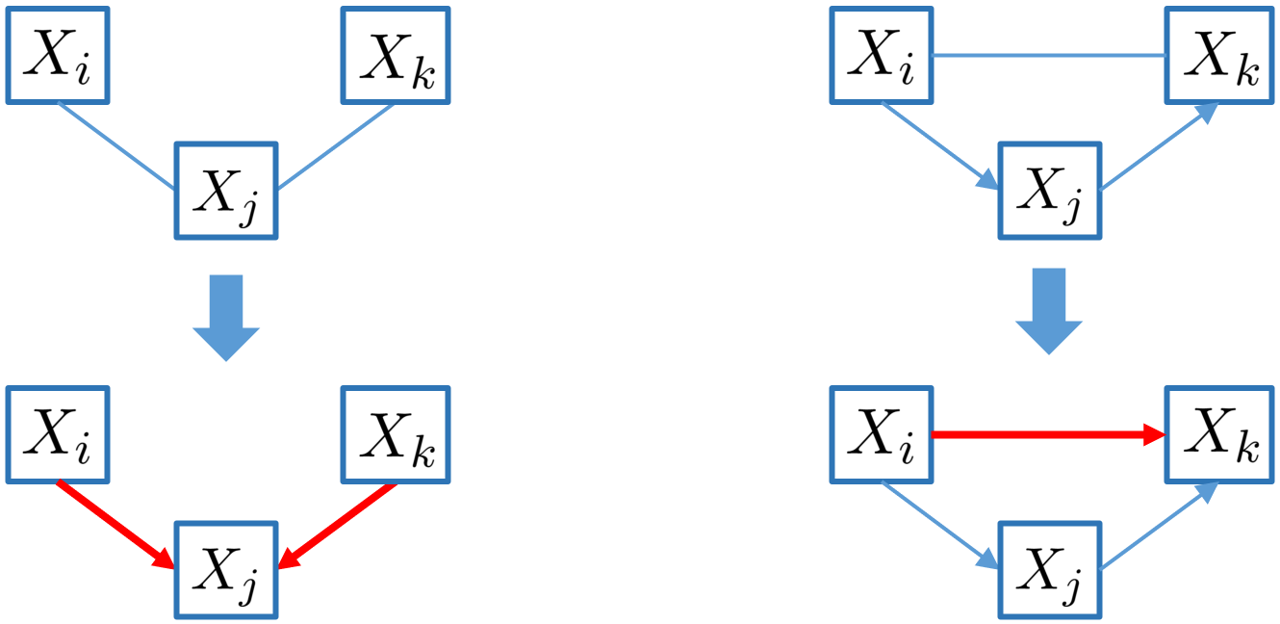}
\caption{Diagram of determining the orientation of edges.}
\label{fig:direction}
\vspace{-1mm}
\end{figure}

The detailed process of using the PC algorithm for Bayesian network structure learning is explained below and also summarized in \textbf{Algorithm~\ref{algorithm:PC}}. The algorithm begins with a complete undirected graph, where the input includes variables $\bm{X}=\{X_1, X_2, ..., X_n\}$. Following that, a conditional independence test is performed on the complete graph over $\bm{X}$. For each connected pair $X_i$ and $X_j$, if the cardinality of $\text{Adj}(C, X_i) \setminus \{X_j\}$ is at least $n$, we check for conditional independence, and remove the edge if conditional independence is confirmed. Repeat the process until the cardinality is less than $n$ for any pair $\{X_i, X_j\}$. 
The algorithm then removes edges from the graph based on conditional independence tests, which leads to an undirected graph called the ``skeleton''. The next step involves determining and updating the orientation of the undirected edges (see Fig.~\ref{fig:direction}). The process is fundamentally based on V-structures and propagation structures, which help prevent cycles and determine the final directions.
Lastly, to enhance the fit for rare-event prediction, modify the learned structure by introducing a directed edge from any node without a child to the target prediction node. As a result, the final outcome is a DAG that captures the conditional dependencies among all the variables in the dataset.

\begin{algorithm}[t!]
% \caption{PC algorithm for Bayesian Network structure learning}
\caption{PC algorithm for Bayesian structure learning}
\label{algorithm:PC}
\begin{algorithmic}[1]
\renewcommand{\algorithmicrequire}{\textbf{Inputs:}}
\renewcommand{\algorithmicensure}{\textbf{Outputs:}}
\REQUIRE Variables $\bm{X}=\{X_1,X_2,...,X_n\}$
\ENSURE A DAG $C$ representing the conditional independence structure of $\bm{X}$

\STATE \textbf{Initialization:} Form a complete undirected graph $C$ where each pair of variables in $\bm{X}$ is connected by an edge.
\STATE \textbf{Initialization:} $n = 0$
\WHILE{there exists an edge between $X_i$ and $X_j$ in $C$ such that $\left|\text{Adj}(C,X_i)\setminus\{X_j\}\right| \geq n$} 
        \FORALL{pairs of variables $(X_i, X_j)$ in $C$}
            \FORALL{subsets $ \bm{S} \subseteq \text{adj}(C,X_i) \setminus \{X_j\} $ with $|\bm{S}|=n$}
                \IF{$X_i, X_j$ are conditionally independent given $\bm{S}$}
                    \STATE Remove edge $X_i-X_j$ from $C$
                \ENDIF
            \ENDFOR
        \ENDFOR
        \STATE $n = n + 1$
    \ENDWHILE
    % \\ $\rightarrow$ $\bigLozenge$ \textbf{Initialization}: \text{Initial feasible solution $\mathbf{x}^{0}$}
    \FORALL{pairs of non-adjacent nodes $(X_i, X_k)$ both adjacent to $X_j$}
        \IF{separating set $\bm{S}$ of $(X_i, X_k)$ does not contain $X_j$}
            \STATE Orient $ X_i-X_j-X_k $ as $ X_i \to X_j \leftarrow X_k $
        \ENDIF
    \ENDFOR
    \FORALL{$ X_i \to X_j \to X_k $}
        \IF{$X_i$ and $X_k$ are adjacent}
            \STATE Orient $X_i-X_k$ as $ X_i \to X_k$
        \ENDIF
    \ENDFOR

    \RETURN The new DAG $C$
\end{algorithmic}
\end{algorithm}

\section{Implementation and Case Studies}
\label{Sec: Case}

With the Bayesian fundamentals and PC structure learning explained, in this section we mainly focus on case studies with realistic datasets to showcase the detailed implementation of the proposed algorithm for predicting transmission line outages.
The case study in this paper is carried out based on the actual transmission line outage data from the Bonneville Power Administration (BPA) \cite{CiteBPAoutage} and historical hourly weather data from the National Oceanic and Atmospheric Administration (NOAA)\cite{CiteNOAAhour}. The Bayesian structure learning and Bayesian network prediction are implemented using Python. All the simulations are performed on a regular laptop with AMD CPU @ 3.30 GHz and 32 GB of RAM.

% \begin{figure}[t!]
% \centerline{\includegraphics[width=6cm]{direction3.png}}
% \vspace{-2mm}
%     \caption{Diagram of determining the orientation of edges.}
%     \label{fig:direction}
%     \vspace{-3mm}
% \end{figure}

During the numerical simulation, we utilize the outage and weather data in Salem, Oregon, USA. The test transmission line group includes 9 lines with operation voltages ranging from 115kV to 230kV. From 1999 to 2023, there were a total of 1925 outages, including 1555 plan outages and 370 auto outages. Among all these outages, 104 were weather-related, with 61 of these being recovered immediately, and other outages occurred at 28 different time points. In this paper, all the analysis is performed on an hourly basis.
\subsection{Prediction Process}
\label{Sec: process}
The detailed process for the prediction of transmission line outage probabilities is summarized as follows:
\begin{enumerate}
  \item \textbf{Data collection and preprocessing}: Collect data from BPA and NOAA. It should be noted that certain factors such as wind direction are excluded in this numerical tests.
  % it should be noted that we may not consider several weather factors this time, such as wind direction. 
  We next use interpolation methods to handle missing data or values from the NOAA data. For the BPA transmission line outage data, we set the corresponding timestamp of the target variable to 1 if a weather-related outage occurs, and to 0 otherwise.
  \item \textbf{Data discretization}: Discrete the historical data for Bayesian network. We discretize each continuous weather factor into 10 bins in this test as an example, and Laplace smoothing method will be adopted if the count of states in a particular bin is zero.
  \item \textbf{Imbalanced data processing}: The test data includes up to 24 years of hourly meteorological data and outage data, while the outage data is relatively rare. To address this imbalance, we first apply down-sampling to the majority class (i.e., normal operation without any outage) to reduce the overall dataset size, and subsequently, we use up-sampling techniques like SMOTE \cite{chawla2002smote} to increase the representation of the minority class ({i.e., weather-induced outages}).
  \item \textbf{Network structure learning}: Apply the PC algorithm for Bayesian structure learning, the specific process of which is shown in Section~\ref{Sec: PC}.
  % \item \textbf{Structure modification}: 
  % Adjust the learned structure to improve its fit for rare-event prediction. In this paper, if a weather factor node lacks child nodes, a directed edge is manually added from the weather factor node to the event node to enhance predictive accuracy.
  \item \textbf{Bayesian network prediction}: Based on the learned structure with modification, apply the Bayesian network for outage probability prediction using real data.
\end{enumerate}

% \subsection{Bayesian Network Prediction}
\subsection{Numerical Validation}
\label{Sec: res}
\begin{figure}[t!]
\vspace{2mm}
\centering
\includegraphics[trim=0cm 0cm 0cm 0cm,clip=true,totalheight=0.14\textheight]{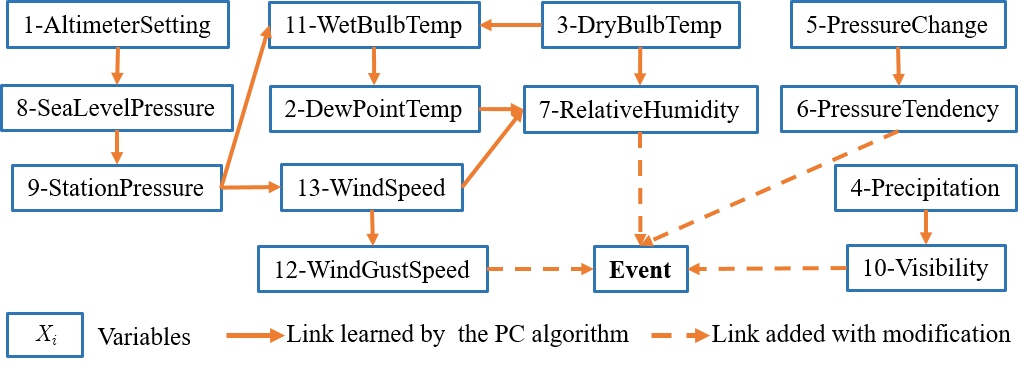}
\caption{PC algorithm results for Bayesian network structure learning.}
\label{fig:PClearn}
\vspace{-1mm}
\end{figure}

Based on the prediction process in Section~\ref{Sec: process}, we test the effectiveness of our method. 
The Bayesian network with the structure learning by the PC algorithm for predicting the outage probability of transmission lines is shown in Fig.~\ref{fig:PClearn}, in which the directed edges in the solid lines represent those learned by the PC algorithm, and the dashed ones are added manually for better prediction.

% \begin{figure}[t!]
% \centerline{\includegraphics[width=8.8cm]{PC structure learning5.png}}
% \vspace{-2mm}
% \caption{PC algorithm results for Bayesian network structure learning.}
% \label{fig:PClearn}
% \vspace{-1mm}
% \end{figure}

Based on the DAG shown in Fig.~\ref{fig:PClearn} and the definition of conditional independence, the value of a specific variable is conditionally dependent only on its parent nodes, and conditionally independent of all non-parent nodes given the value of its parent nodes. Therefore, the conditional probability of the outage probability can be expressed as:
\begin{equation}
\begin{split}
P(E \mid \bm{F}) = P(E \mid F_6,F_7,F_{10},F_{12})
\label{eq:dag}
\end{split}
\end{equation}
Here, $E$ denotes the event of transmission lines outage, $\bm{F}$ is the state of all meteorological factors, and $F_i$ is the state of meteorological factor $i$.
Therefore, $F_6,F_7,F_{10},F_{12}$ represent pressure tendency, relative humidity, visibility, and wind gust speed, respectively (see Fig.~\ref{fig:PClearn}).

Using Bayes' theorem, \eqref{eq:dag} can be further formulated as:
\begin{equation}
\begin{split}
\textstyle{P(E \mid F_6,F_7,F_{10},F_{12}) = \frac{P(E \cap F_6 \cap F_7 \cap F_{10} \cap F_{12})}{P(F_6,F_7,F_{10},F_{12})} }
\label{eq:dag2}
\end{split}
\end{equation}
Considering the Markov property of the Bayesian network, the factorization of the two joint distributions in \eqref{eq:dag2} can be given as the following:
\begin{equation}
\begin{split}
P(E \cap F_6 \cap F_7 \cap F_{10} \cap F_{12})=P(F_6)\cdot P(F_7\mid F_6)\\
\cdot P(F_{10}\mid F_6,F_7)
\cdot P(F_{12}\mid F_6,F_7,F_{10})\\
\cdot P(E\mid F_6,F_7,F_{10},F_{12})
\label{eq:dag up}
\end{split}
\end{equation}
\begin{equation}
\begin{split}
P(F_6,F_7,F_{10},F_{12})=
P(F_6 \mid F_5)\cdot 
P(F_7 \mid F_5, F_3, F_{13}) \\ 
\cdot
P(F_{10} \mid F_4) \cdot
P(F_{12} \mid F_{13})
\label{eq:dag down}
\end{split}
\end{equation}

\begin{figure}[t!]
\centerline{\includegraphics[width=8cm]{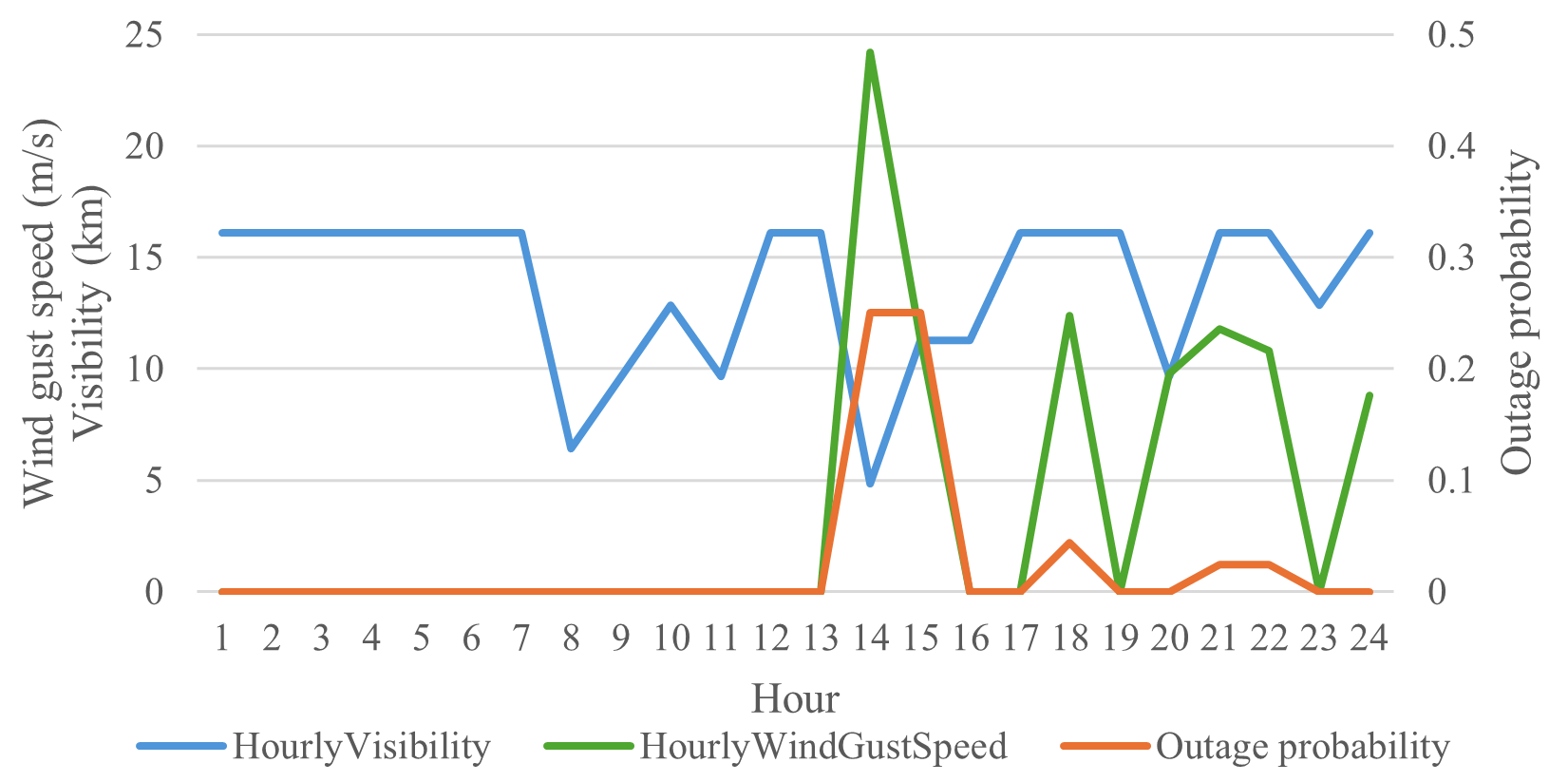}}
\vspace{-1mm}
\caption{Bayesian prediction results using PC algorithm for structure learning.}
\label{fig:BNresults}
% \vspace{-1mm}
\end{figure}

Accordingly, the conditional outage probability of transmission lines under specific weather conditions can be calculated using maximum likelihood estimation.
% \textcolor{red}{Based on the learned structure and parameter learning, if there is missing data, which is the parent node of the event, we can still calculate the outage probability. ???}
% Given the limited outage records, we analyze a group of transmission lines collectively. Therefore, the conditional outage probability $P(E\mid \bm{F})$ reflects the outage probability for the entire line group. Assuming every line within this group has the same outage probability, the outage probability for each line can be given as:
% \begin{equation}
% \begin{split}
% (1-P(e\mid \bm{F}))^n=1-P(E \mid \bm{F})
% \label{eqs1}
% \end{split}
% \end{equation}
% where $P(e\mid \bm{F})$ is the conditional outage probability of each line under weather condition $\bm{F}$, and $n$ is the total number of lines in the group.
The prediction results of our proposed method are shown in Fig.~\ref{fig:BNresults}. A single day is selected for illustration with an outage occurring at around 14:48. 
% All weather factors are considered in the prediction, while only wind gust speed and visibility are shown in the figure.
While all the meteorological factors have been considered for the outage prediction, for visualization and demonstration purposes, we specifically select two meteorological factors: wind gust speed and visibility. Results show that at 14:00 and 15:00, the predicted outage probabilities are much higher than normal levels, and thus preventive actions are recommended for implementation before this period. 

% For weather conditions with outages, the average outage probability of the line group is 0.4478, and the outage probability of each line in the group is 0.3693.

\begin{table}[t!]
\caption{Comparison of different outage prediction methods}
\begin{center}
\centerline{\includegraphics[width=7.5cm]{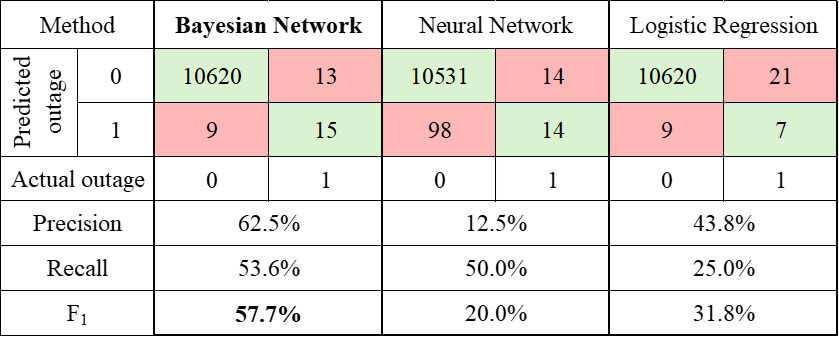}}
\label{tab:Compare}
\end{center}
% \vspace{-2mm}
\end{table}

\subsection{Comparison Studies}
\label{Sec: com}
For weather-related line outage predictions, there are plenty of machine learning algorithms that are applicable. Out of all these available algorithms, in this work, we select the more commonly used neural network (NN) and logistic regression (LR) to compare with the proposed Bayesian network method.
% In existing research, neural network (NN) and logistic regression (LR) methods are commonly used for outage prediction. 
To demonstrate the advantages of our proposed method, we will compare its performance with both NN and LR. 
For clearer comparison, we first select several metrics in classification problems for evaluation: True Positives (TP) are events correctly predicted as positive, True Negatives (TN) are correctly predicted as negative, False Positives (FP) are events incorrectly predicted as positive when they were actually negative, and False Negatives (FN) are events incorrectly predicted as negative when they were actually positive.

For transmission line outage probability prediction, considering the risks of power outages and the costs of preventive measures, we will focus on the metrics of TP and FP. Therefore, precision, recall, and F$_1$-score are also selected as evaluation metrics.
% ${\cal{F}}_1$
Specifically, precision is defined as the proportion of TP to the amount of total positives that the model predicts. 
\begin{equation}
\text{Precision}=\frac{\text{TP}}{\text{TP}+\text{FP}}
\label{eq:precision}
\end{equation}

Recall focuses on how good the model is at finding all the positives. 
\begin{equation}
\text{Recall}=\frac{\text{TP}}{\text{TP}+\text{FN}}
\label{eq:recall}
\end{equation}

F$_1$-score is the harmonic mean of the precision and recall, which reflects a strong performance in recognizing positive cases while minimizing FP and FN. It is particularly useful for evaluating balanced performance, especially in predictions with imbalanced data.
% and it could also indicate a well-balanced performance, especially for predictions with imbalanced data. 
\begin{equation}
\text{F}_1\text{-score}=2\cdot \frac{\text{Precision} \cdot \text{Recall}}{\text{Precision}+\text{Recall}}
\label{eq:F1}
\end{equation}

\begin{figure}[t!]
\centerline{\includegraphics[width=7.5cm]{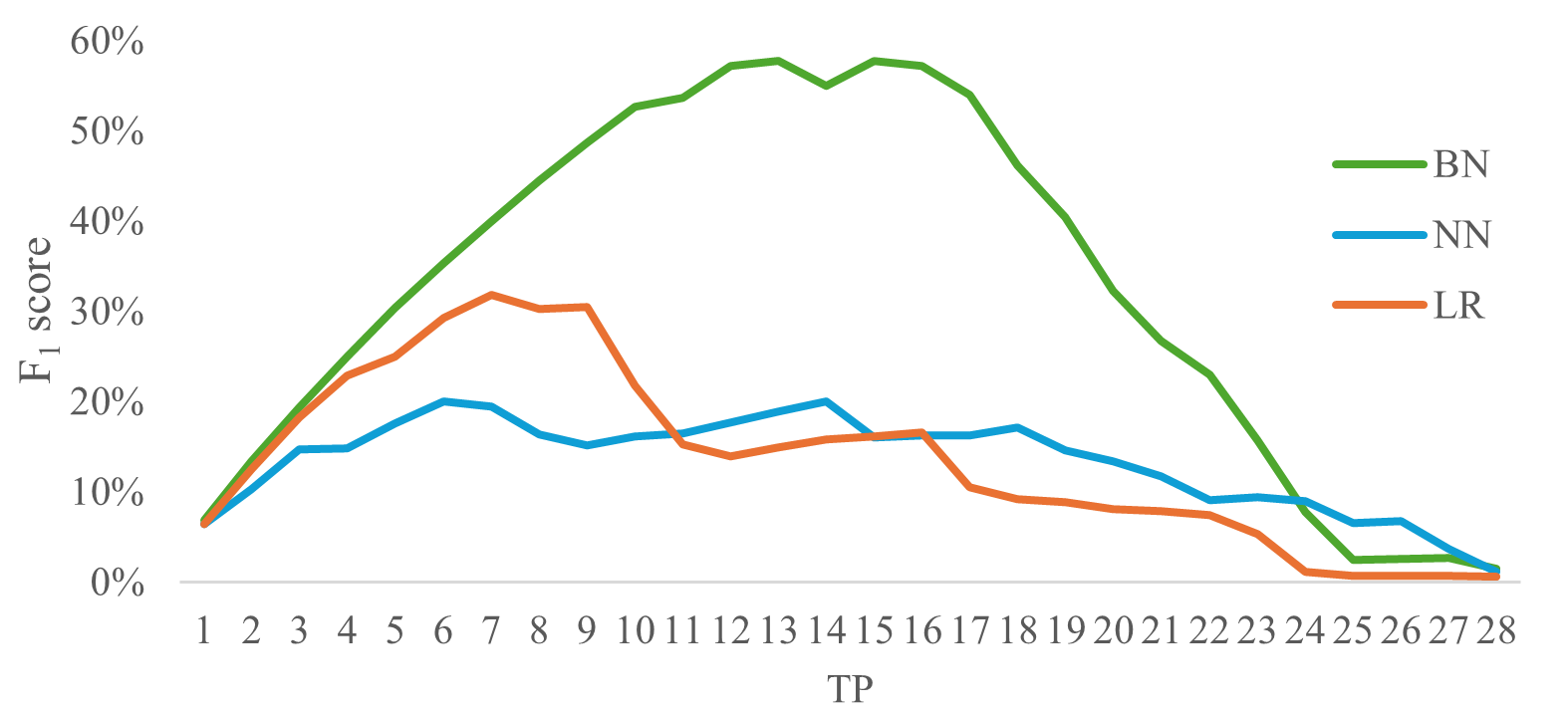}}
\vspace{-2mm}
\caption{Comparison of F$_1$-scores different outage prediction methods.}
\label{fig:F1 score}
% \vspace{-2mm}
\end{figure}

We randomly select 5\% of the data for algorithm validation, while ensuring that all outage records are included in the validation dataset due to the rarity of outages.
% About 5\% of all records are selected for model validation, and it should be noted that all outage records are included in the validation dataset considering the rarity of outages. 
The performance comparisons of different outage prediction methods with the best F$_1$-score are shown in TABLE~\ref{tab:Compare}.  It is clear that at the highest F$_1$-score for each method, our proposed method demonstrates the best performance across precision, recall, and F$_1$-score metrics. In addition, the overall performance indicated by F$_1$-score, is demonstrated in Fig.~\ref{fig:F1 score}. our proposed method achieves the highest score in most cases, which indicates that our method has good performance in both precision and recall rate.

\begin{table}[t!]
\caption{Comparison of different methods with imbalanced data}
\begin{center}
\centerline{\includegraphics[width=7.5cm]{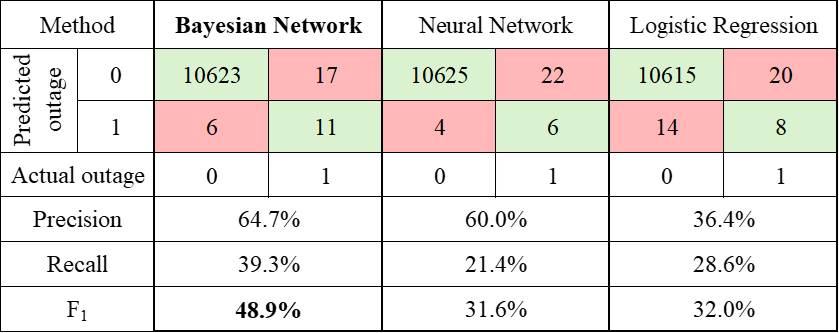}}
\label{tab:Compare-imbalanced}
\end{center}
\vspace{-7mm}
\end{table}

% \begin{figure}[t!]
% \centerline{\includegraphics[width=8cm]{F1 score5.png}}
% \caption{Comparison of different outage prediction methods}
% \label{F1 score}
% \end{figure}

To further verify the robustness of the algorithm, we also validate the performance of these methods using imbalanced data. That is, the up-sampling process is not applied for learning, training or regression datasets in the test, and the corresponding results are provided in TABLE~\ref{tab:Compare-imbalanced}. 
Our method also outperforms other methods, which highlights its robustness under different scenarios. By improving the outage prediction performance, the proposed Bayesian learning algorithm can help utilities effectively reduce resources required for preventive measurements and the operational costs when responding to the predicted outages.

\section{Conclusions and Future Work}
\label{Sec: conclu}

Climate change has caused an increasing number of severe weather events in recent years, which poses threats to the resilient operation of power grids. To better prepare the grids for potential emergencies, it is critical to develop an accurate probability model for these weather-related power outages. Traditional methods for line outage predictions typically offer a risk level but lack the ability to provide precise probabilities. In this work, we propose a novel Peter-Clark (PC) structural learning-based Bayesian network to address this limitation. In addition, the PC structural learning design enhances the efficiency and scalability in obtaining the Bayesian network, allowing it to perform effectively even with limited outage data. Case studies using realistic data from BPA and NOAA have demonstrated the effectiveness of the proposed algorithm, and comparisons with several existing methods further highlight its advantages. Future research includes applying the proposed method and prediction results to system-level resilience-related operation and control problems. To that end, we plan to also incorporate economic losses and social equity aspects during extreme events into the future studies.

% \section*{Acknowledgment}

% Thanks to XXX 

% \textcolor{red}{discuss impacts (complexity)}

\vfill

\bibliography{bibliography.bib}

\bibliographystyle{IEEEtran}

\end{document}